%% file: main.tex
\documentclass[10pt,twocolumn,letterpaper]{article}

\usepackage{cvpr}
\usepackage{times}
\usepackage{epsfig}
\usepackage{graphicx}
\usepackage{amsmath}
\usepackage{amssymb}
\usepackage{color}
\usepackage{latex_pkgs/maths}
\usepackage{latex_pkgs/msformatting}
\usepackage{algorithm2e}
\usepackage{listings}
\usepackage{cuted} % for strip

\usepackage[pagebackref=true,breaklinks=true,letterpaper=true,colorlinks,bookmarks=false]{hyperref}

\cvprfinalcopy % *** Uncomment this line for the final submission

\def\cvprPaperID{****} % *** Enter the CVPR Paper ID here

\ifcvprfinal\fi
\begin{document}

%%%%%%%%% TITLE
\title{Action Detection from a Robot-Car Perspective}

\author{Valentina Fontana, Manuele Di Maio\\
Universit\'a degli Studi Federico II\\
Naples, Italy\\
{\tt\small vale.fontana@studenti.unina.it, man.dimaio@gmail.com}
\and
Stephen Akrigg, Gurkirt Singh, Suman Saha, Fabio Cuzzolin\\
Oxford Brookes University\\
Oxford, UK\\
{\tt\small 15057204@brookes.ac.uk, gurkirt.singh-2015@brookes.ac.uk,} \\ {\tt\small suman.saha-2014@brookes.ac.uk, fabio.cuzzolin@brookes.ac.uk}
}

\maketitle
%\thispagestyle{empty}

%%%%%%%%% ABSTRACT

% \newcommand{\red}[1]{\textcolor{red}{#1}}
% \newcommand{\GUR}[1]{\textcolor{blue}{#1}}
% \newcommand{\SAHA}[1]{\textcolor{red}{#1}}

\begin{abstract}  
\input{text/abstract}
%\vskip -1cm
\end{abstract}
\input{text/intro}
% \input{text/contrib}
\input{text/soa}
\input{text/robot-car-dataset}
\input{text/action-detection-methods}
\input{text/experiments}
\input{text/conclusion}
%%%%%%%%%%%%%%%%%%%%%%%%%%%%%%%%%%%%%%%%%%%%%%
% \clearpage
{\small
\bibliographystyle{ieee}
\bibliography{bib}
}
\end{document}

%% file: text/abstract.tex
We present the new Road Event and Activity Detection (READ) dataset, designed and created from an autonomous vehicle perspective to take action detection challenges to autonomous driving. READ will give scholars in computer vision, smart cars and machine learning at large the opportunity to conduct
research into exciting new problems such as understanding complex (road) activities, discerning the behaviour of sentient agents, and predicting both the label and the location of future actions and events, with the final goal of supporting autonomous decision making.
%Unlike current spatiotemporal human action detection dataset,  READ deals with the actions for all types of road users like cyclists, motor-bikers, small/large vehicles and pedestrians, not just humans.

%Even though humans perform cars actions implicitly, but a robot-car only see the cars doing those actions. 
%And, unlike other  

%% file: text/intro.tex
\section{Introduction} \label{sec:intro}
% punchline
With a rapid increase in the number of cars and other vehicles in the urban transportation system, 
\emph{autonomous driving} (or \emph{robot-assisted driving}) has emerged as one of the predominant research areas in artificial intelligence. 
% some motivating examples
Imagine a self-driving car allowing you to catch a bit of sleep while you are on your way to office/school, 
watch a movie with your family on a long road trip or drive back home after a night out at the bar.
% GUR: Maybe mention kitti and Stanley as well rather than attributing the whole to GOOGle and Toyota 
Work towards the development of such advanced autonomous cars has dramatically increased since the achievements of Stanley
\cite{thrun2006stanley} in the 2005 Darpa grand challenge~\cite{buehler20072005}. 
In recent years many large companies such as Toyota, Ford, Google have introduced their own versions of the robot car concept~\cite{KirstenNov2017,GoogleWAYMO,pandey2011ford}. 
As a result, ``self-driving cars'' are increasingly considered to be the next big step in the development of personal use vehicles.
\\
Society's smooth acceptance of this new technology, however, depends on many factors such as safety, ethics, cost and reliability, to name a few. 
For example, from a safety perspective in a mixed scenario in which both robots and humans share the road, smart cars need to be able to spot children approaching 
a zebra crossing and pre-emptively adjust speed 
and course to cope with the children’s possible decision to cross the road.
At the same time, their cost should be affordable for the average consumer.
At present, though, the vast majority of these cars do not meet 
all these predefined standards to make them available to the public.

The latest generation of robot cars use a range of different 
sensors (i.e. laser rangefinders, radar, cameras, GPS) to provide data on what happens on the road, and fuse the information extracted from all these modalities in a meaningful 
way to suggest how the car should maneuver~\cite{GoogleWAYMO}.
\\
% \hlb{we need to mention autonoumus car dataset and what they lack; then introduce what we propose}
A number of autonomous car datasets exist for 3D environment mapping~\cite{pandey2011ford}, 
stereo reconstruction~\cite{pfeiffer2013exploiting}, 
or both~\cite{blanco2014malaga}, including 
optical flow estimation and self localisation~\cite{geiger2013vision}. 
Recently, Maddern \etal introduced a large scale dataset~\cite{maddern20171} 
for self localisation via LIDAR and vision sensors. 
All these benchmarks are designed to address interesting problems -- nevertheless, none of them tackles the paramount issue of allowing the car to be aware of the actions performed by surrounding vehicles and humans, and in general of detecting, recognising and anticipating complex road events to support autonomous decision making.

Thus, in this paper, consider the issue of vision-based autonomous driving, i.e., the problem of endowing cars to self-drive based on streaming videos captured by cameras mounted on them. 
In such a setting, which closely mimicks how human drivers `work', the car needs to reconstruct and understand the surrounding environment from the incoming video sequence(s). A crucial task of video understanding is to recognise and localise (in space and time) different actions or events appearing in the video: for instance, the vehicle needs to perceive the behaviour of pedestrians by identifying which kind of activities (e.g., `moving' versus `stopping') they are performing, when and where~\cite{sahaphdthesis2018} this is happening.
In the computer vision literature \cite{Georgia-2015a,Weinzaepfel-2015,saha2017amtnet,kalogeiton2017action,hou2017tube,Saha2016,peng2016eccv,singh2016online}
this problem is termed \emph{spatio-temporal action localisation} or, in short, \emph{action detection}.
Although a considerable amount of research has been undertaken in this area,
most approaches perform offline video processing and are thus not suitable for self-driving cars which require the online processing of streaming video frames at real-time speed. 
In opposition, most recently Singh~\etal~\cite{singh2016online} have proposed an online, real-time action detection approach. However, as it is common practice in the action detection community, they evaluated their model on action detection datasets composed by YouTube video clips not designed from a robot car perspective.

Unlike current human action detection 
datasets \cite{UT-Interaction-Data} such as J-HMDB~\cite{J-HMDB-Jhuang-2013}, UCF-101 \cite{soomro-2012}, LIRIS-HARL \cite{liris-harl-2014}, DALY \cite{daly2016weinzaepfel} or AVA \cite{ava2017gu}, 
the Road Event and Activity Detection (READ) dataset
we introduce here is specially designed from the perspective of self-driving cars, and includes spatiotemporal actions performed not just by humans but by all road users, including cyclists, 
motor-bikers, drivers of vehicles large and small, and obviously pedestrians. 
\\
We strongly believe, a belief back up by clear evidence, that an awareness of all the actions and events taking place, and their location within the road scene, is essential for inherently safe self-driving cars. To this purpose we introduce three different types of label for each such road event, namely: 
(i) the position of the road user relative the autonomous vehicle perceiving the scene (e.g. in vehicle lane, on right pavement, in incoming lane, in outgoing lane); (ii) the type of the road user (e.g. pedestrian, small/large vehicle, cyclist); and (iii) the type of action being performed by the road user (e.g. moving away, moving towards, crossing the road, crossing the road illegally, and so on). 
%GUR maybe we should mention here how it is different from current action datasets
\\
READ has been generated by providing additional annotation for a number of videos captured by the cameras mounted on the Oxford RobotCar platform, an autonomous Nissan LEAF, while driving in the streets of the city of Oxford, in the United Kingdom (\S~\ref{sec:dataset}). All such videos are part of the publicly available Oxford RobotCar Dataset~\cite{maddern20171} released in 2017 by the Oxford Robotics Institute\footnote{\url{http://robotcar-dataset.robots.ox.ac.uk/}}. More specifically, ground truth labels and bounding box annotations (which indicate where the action/event of interest is taking place in each video frame) are provided for several actions/events taking place in the surroundings of the robot-car (\S~\ref{sec:dataset}).

To the best of our knowledge, READ is the first action detection dataset 
which can be fully exploited to train machine learning algorithms tailored for self-driving robot cars. Additionally, it significantly expands the range and scope of current action detection benchmarks, in terms of size, context, and specific challenges associated with the road scenario.
Here we report quantitative action detection results produced on READ by what is currently the state-of-the-art online action detection 
approach~\cite{singh2016online}.

%Further down the line, as a next step in the progression of action detection from robot car perspective, we outline a deep network architecture for detection of complex human activities based on a deep net implementation of deformable part based models~\cite{girshick2015deformable}.

%% file: text/soa.tex
\section{Related work}

Most of current generation visual datasets for autonomous driving address issues like 3D environment mapping~\cite{pandey2011ford} or
stereo reconstruction~\cite{pfeiffer2013exploitingblanco2014malaga}. 
A large-scale dataset called KITTI was released in 2013 for optical flow estimation and self localisation~\cite{geiger2013vision}. 
Similarly, Maddern \etal introduced in 2017 a large scale dataset~\cite{maddern20171} 
for self localisation via LIDAR and vision sensors. 
The relevant effort closest to the scope of READ is due to
Ramanishka~\etal~\cite{ramanishka2018toward}, who deal with action and events in the car context. The authors, however, limit themselves to the behaviour of the driver rather than looking at events involving other cars.
As stated above, we think, instead, that a full awareness of events or activities performed by other road users is necessary for an autonomous car to successfully navigate complex road situations. As a consequence, READ considers the problem of detecting road events and activities performed by other road users as well as the robot-car itself.  

Inspired by the record-breaking performance of CNN-based 
object detectors~\cite{redmon2016yolo9000,ren2015faster,liu15ssd} several scholars
~\cite{singh2016online,Saha2016,Georgia-2015a,peng2016eccv,Weinzaepfel-2015,weinzaepfel2016towards,zolfaghari2017chained} 
have recently extended object detectors to videos for spatio-temporal action localisation. This includes, in particular, a recent work by Yang \etal \cite{yang2017spatio} which uses features extracted from the current, frame $t$ proposals to `anticipate' region proposal locations at time $t+\Delta$ and use them to generate future detections. None of these approaches, however,  
tackle spatial and temporal reasoning jointly at the network level, 
as spatial detection and temporal association are treated as two disjoint problems.
More recent efforts try to address this problem by predicting `micro-tubes'~\cite{saha2017amtnet} or,
alternatively, `tubelets'~\cite{kalogeiton2017action,hou2017tube}, for sets of frames taken together.
However, to be applicable to the road event detection scenario, these methods need to run in real-time and in an online fashion -- 
for this reason, here we select~\cite{singh2016online} by Singh~\etal
as a baseline to conduct experiments on READ.

%% file: text/robot-car-dataset.tex
\section{Road Event and Activity Detection dataset} \label{sec:dataset}

There are six cameras on Oxford RobotCar~\cite{maddern20171}, where three front-facing cameras are for the stereo generation. 
In our annotation process, we use videos captured from the central camera from the stereo setup. 
The desired annotations are produced in four steps, as explained in the following subsections.

\subsection{Multi-label concept}

We consider all the possible agents, the actions they perform, their locations with respect to the `autonomous vehicle' (AV) and the actions of the vehicle itself. 
Multiple agents may be present at any given time, and perform multiple actions simultaneously. 
We propose to label each agent using at least one label, and locate its position using a bounding box around it. % Fabio: show an example frame??

% All the agets carry same agent label throughout the scence expect traffic lights can change labels e.g. light can change from red to grren.

\subsubsection{Agent labels}

We consider three types of actors as main road users, as well as traffic lights as a class of object that can perform actions able to influence the decision of AV.
In READ we call them \emph{agents}. AV is considered as just another agent. 
These three agent classes are: pedestrian, vehicle and cyclist. 
Further, the vehicle category is subdivided into six sub-classes: two-wheeler, car, bus, small-size vehicle, medium-size vehicle, large-size vehicle. 
Similarly, the `traffic light' agent class is subdivided into two sub-classes (Table~\ref{table:objects}), one referring to traffic lights in the 
AV lane and the other to vehicles in a different lane. 
Each traffic light class can be associated to three action classes: red, amber and green (see Table \ref{table:actions}).
%A full list of agents labels is given in table~\ref{table:objects}. 
Only one out of the 11 agent labels in Table~\ref{table:objects} can be assigned to each agent present in the scene.

\begin{table}[h]
  %\vskip -3mm
  \centering
  \setlength{\tabcolsep}{4pt}
  \caption{Agent labels.}
  \vspace{2mm}
  {\footnotesize
  \scalebox{1}
  {
  \begin{tabular}{c}
  \toprule
    Autonomous Vehicle (AV)\\
    Cyclist\\
    Pedestrian\\
    Car\\
    Bus\\
    Two-wheeler\\
    Small vehicle\\
    Medium vehicle\\
    Large vehicle\\
    Vehicle traffic light \\
    Other traffic light \\
%     Vehicle traffic light red\\
%     Vehicle traffic light amber\\
%     Vehicle traffic light green\\
%     Other traffic light red\\
%     Other traffic light amber\\
%     Other traffic light green\\
  \bottomrule
  \end{tabular}
  }
  }
  \label{table:objects} %\vspace{-6mm}
\end{table}

\begin{figure*}[t]
  \centering
  \includegraphics[scale=0.8]{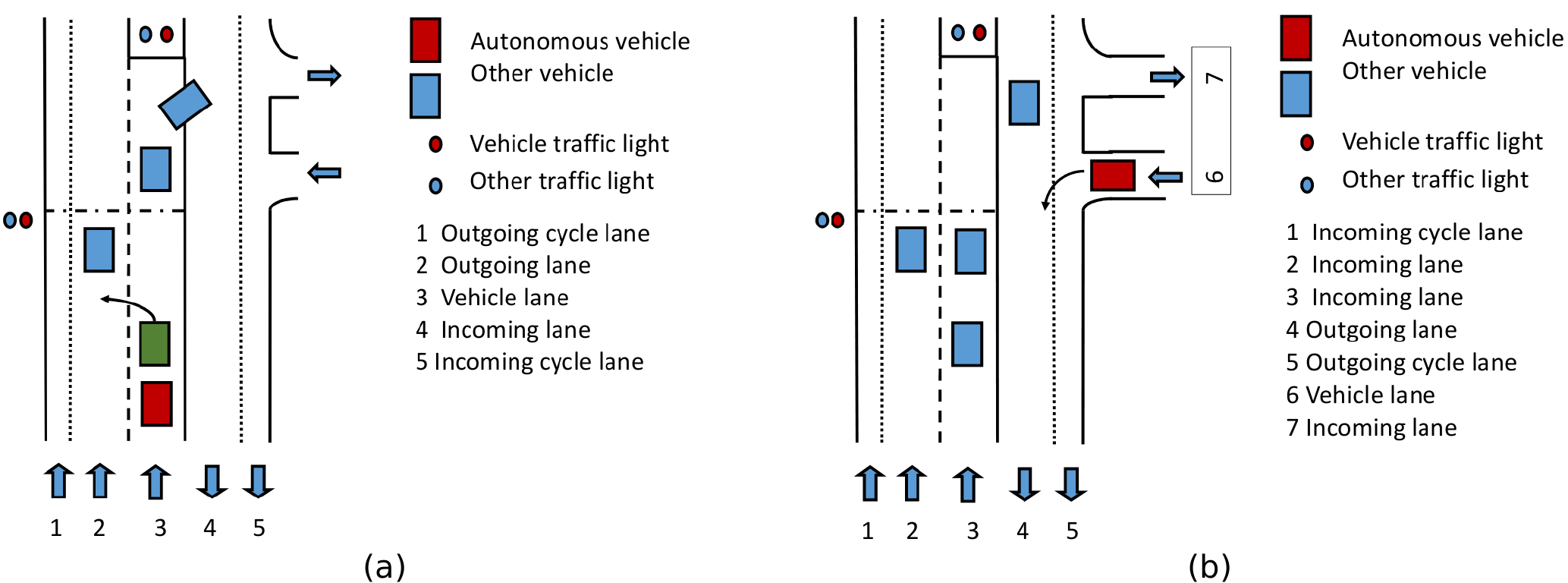}
  %\vskip 2mm
  \caption{
    {\small
      \textit{Illustration of location labelling. Sub-figure (a) shows a green car in front of the Autonomous Vehicle changing lanes, as depicted by the arrow symbol. The associated event will then carry the following labels: `In vehicle lane', `Moving left', `Merging'. Once the merging action is  completed, the location label changes to `In outgoing lane'. In sub-figure (b), if the Autonomous vehicle is to turn left from lane 6 to lane 4, then lane 4 is the `outgoing Lane' as the traffic is moving in the same direction as the AV, will be once it completes its turn. However, if the Autonomous vehicle is to turn right from lane 6 to lane 4 (a wrong turn), then lane 4 will be the `incoming lane' as the vehicle will be moving into the incoming traffic.}
    }
 }
  \label{figure:locations} %\vspace{-3mm}
\end{figure*}

\begin{table}[h]
  %\vskip -3mm
  \centering
  \setlength{\tabcolsep}{4pt}
  \caption{Action Labels.}
  \vspace{2mm}
  {\footnotesize
  \scalebox{1}
  {
  \begin{tabular}{c}
  \toprule
    Moving \\
    Moving away\\
    Moving towards\\
    Revering\\
    Breaking\\
    Stopped\\
    Indicating left\\
    Indicating right\\
    Hazard lights on\\
    Looking behind\\
    Turning left\\
    Turning right\\
    Moving right\\
    Moving left\\
    Merging \\
    Overtaking road user\\
    Waiting to cross\\
    Crossing road from left\\
    Crossing road from right\\
    Pushing object\\
    Traffic light red\\
    Traffic light amber\\
    Traffic light green\\
  \bottomrule
  \end{tabular}
  }
  }
  \label{table:actions} %\vspace{-6mm}
\end{table}

\subsubsection{Action labels}
% There can only  one actor label per object or actor. 
% Each object or actor can have only one label assigned to it.
% However, there can be multiple to none action label per actors, e.g. traffic lights carry only object label, but a car carries turning right and indicating right as two labels at the same time. 

Each agent can carry one or more labels at any given time instant.
For example, a traffic light can only carry a single action label - either red, amber or green, whereas a car can be associated with two action labels simultaneously, e.g., `turning right' and `indicating right'.

Although some road agents are inherently multi-tasking, %nature is inherent in modern-day road users, but 
some multi-tasking combinations can be suitably described by a single label e.g. 
`pushing a trolley while walking on the footpath' can be simply labelled as `pushing a trolley'. 
We list all the action labels considered in READ in Table~\ref{table:actions}.
\\
\noindent
\textbf{AV actions.}
Each video frame is also labelled with the action label associated with the AV. In order to accomplish this, a bounding box is drawn on the bonnet of the AV and labelled. 
We assign to the AV one of the six action labels: `moving', `stopped', `turning left', `turning right', `merging', `overtaking road user') . 
These labels are similar to those used for the AV in~\cite{ramanishka2018toward}, while being of a more abstract nature. In addition, as explained, READ covers many more events and actions, performed by other vehicles. 

\subsubsection{Agent location labels}

Agent location is crucial in deciding what action the AV should take next.
% essential in deciding type of action AV should take. 
As the final objective is to assist autonomous decision making, we propose to label the location of each agent from the perspective of the AV.
To understand this, Figure~\ref{figure:locations} illustrates two scenarios in which the location of the other vehicles sharing the road is depicted from the point of view of the AV.
Table~\ref{table:locations} shows all the possible locations an agent can assume, e.g., a pedestrian can be on the right or the left pavement, or in vehicle lane, or at the crossing or at a bus stop.
The same applies to other vehicles as well. There is no location label for the traffic lights as they are not movable objects, but agents of a static nature.

\begin{table}[h]
  %\vskip -3mm
  \centering
  \setlength{\tabcolsep}{4pt}
  \caption{Location Labels.}
  \vspace{2mm}
  {\footnotesize
  \scalebox{1}
  {
  \begin{tabular}{c}
  \toprule
    In outgoing bus lane\\
    In incoming bus lane\\
    In outgoing cycle lane\\
    In incoming cycle lane\\
    In vehicle lane\\
    In outgoing lane\\
    In incoming lane\\
    On left pavement\\
    On right pavement\\
    At junction\\
    At traffic lights\\
    At crossing\\
    At bus stop\\
  \bottomrule
  \end{tabular}
  }
  }
  \label{table:locations} %\vspace{-6mm}
\end{table}

\subsection{Annotation process}

\subsubsection{Video collection}

In the setup by Maddern~\etal~\cite{maddern20171} there are three front-facing cameras.
As part of READ we only annotate the video sequences recorded by the central camera, downloaded from the Oxford RobotCar dataset website (\url{http://robotcar-dataset.robots.ox.ac.uk/}).
Image sequences are first demosaiced to convert them into RGB image sequences and then encoded into video sequences using ffmpeg\footnote{\url{https://www.ffmpeg.org/}} at the rate of 12 frames per second (fps). 
Although the original frame rate in the considered image sequences varies from 11 fps to 16 fps, we uniformised it to keep the annotation process consistent.
Only a subset of all the available videos was selected based on content, by manually inspecting the videos. %Fabio: can we provide the exact list of videos? prob in the final version
Annotators were asked to select videos in order to cover all types of labels and avoid heavy tail as much as possible.

\subsubsection{Annotation tool}

Annotating tens of thousand of frames rich in content is a very intensive process, and calls for a tool which is both fast and user-friendly
\\
After trying multiple tools, such as the Matlab Autonomous Driving System toolbox\footnote{\url{https://uk.mathworks.com/products/automated-driving.html}} and Vatic, we decided to use an open source tool available on the GitHub by Microsoft, called Visual Object Tagging Tool (VOTT, \url{https://github.com/Microsoft/VoTT}). 
The most useful feature of VoTT is that it can copy annotations (bounding boxes and their labels) from the previous frame to the current frame, so that boxes across frames are automatically linked together. 
We used the most basic version of the VOTT without any detector or tracking. 
However, the annotation copy (from the previous frame) property of VOTT allows us to link the bounding boxes across time implicitly. 
VOTT also allows for multiple labels, as in our multi-label annotation concept which requires us to label location, agent and action simultaneously. 
%A single annotator does all the annotation in a video without any verification from any other annotator.

\subsection{Final event label creation}\label{ssc:event_labels}

Given annotations for actions and agents in the multi-label scenario as discussed above, 
we can generate event-level labels pertaining to the agents, 
e.g. \textit{`pedestrian moving towards the AV on the right pavement'},
\textit{`cyclist overtaking in the vehicle lane'} etc.
These labels can be any combinations of location, action and actor labels. If we ignore the location labels the resulting event labels become location invariant. 
When creating event-level labels a trade-off needs to be struck,
depending on the final application and the number of instances available in the dataset for a particular event. 

\begin{figure*}[t]
  \centering
  \includegraphics[scale=0.19]{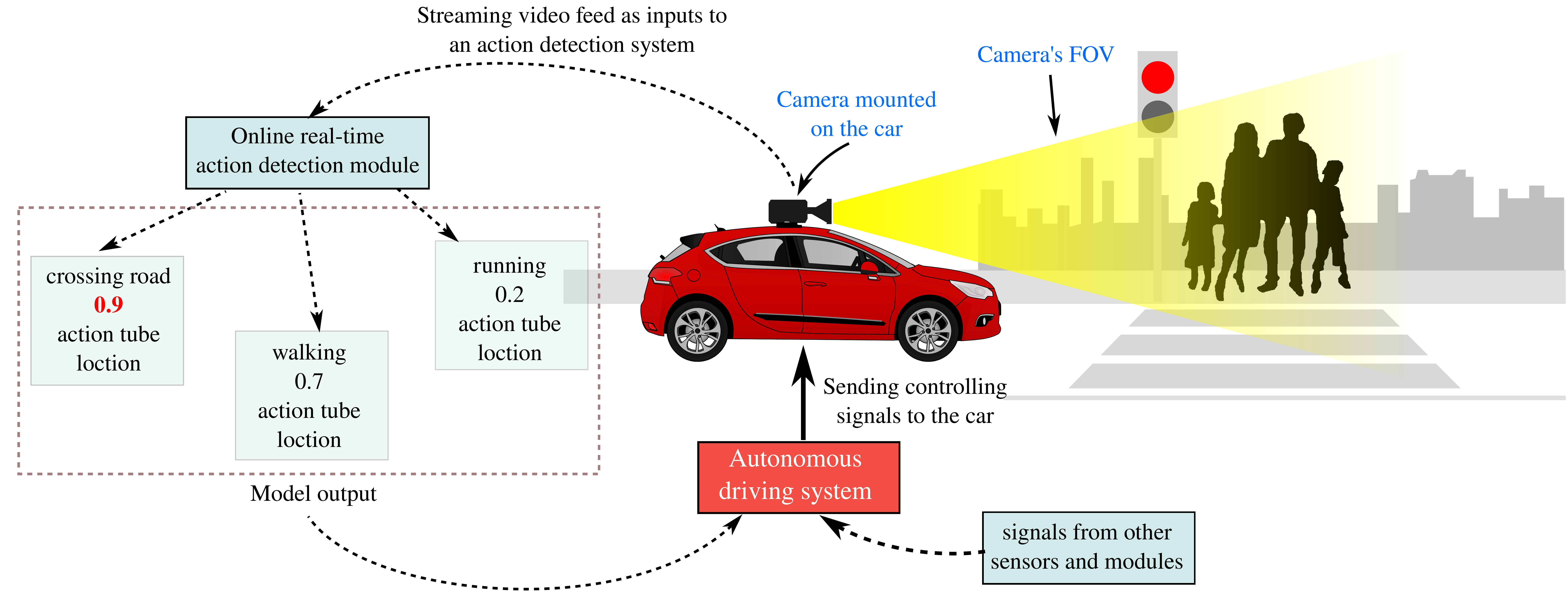}
%   \vskip 2mm
  \caption{{\small \textit{An ideal autonomous driving system in action.}}}
 
\label{fig:current_methods} 
% \vspace{-0.6cm}
\end{figure*}

\subsection{Complex activity label creation}

The purpose of READ is to go beyong the detection of simple actions (as is typical of current action detection datasets), to provide a benchmark test-bed for \emph{complex road activities}, defined as ensembles of events and actions performed by more than one agent in a correlated/coordinated way.
A complex activity is thus made up of simple actions. E.g., `Illegal crossing' 
is composed of `Pedestrian crossing road' + 
`Vehicle traffic light green' + `Vehicle braking' + 
`Vehicle stopped at traffic light'. \\
%It can be understood and activity by an agent in the context of a particular context, e.g. traffic light indication or road marking. 
A list of READ complex activities is shown in Table~\ref{table:complexactivities}.

\begin{table}[h]
  %\vskip -3mm
  \centering
  \setlength{\tabcolsep}{4pt}
  \caption{Complex activity labels.}
  \vspace{2mm}
  {\footnotesize
  \scalebox{1}
  {
  \begin{tabular}{c}
  \toprule
Pedestrian crossing road legally\\
Pedestrian crossing road illegally\\
Vehicle stopping at traffic light\\
Vehicle stopping for crossing\\
Vehicle doing three point turn\\
Vehicle doing U-turn\\
Bus stopping at the bus stop\\
Vehicle parking\\
Cyclist riding legally\\
Cyclist riding illegally\\
Vehicle avoiding collision by slowing down\\
Vehicle avoiding collision by moving to another lane \\
Vehicle stopping temporary\\
Vehicle indicating hazard manoeuvre\\
Vehicle moving for emergency vehicle\\
Vehicle stopping for emergency vehicle\\
Vehicle avoiding a stationary object\\
  \bottomrule
  \end{tabular}
  }
  }
  \label{table:complexactivities} %\vspace{-6mm}
\end{table}

%% file: text/action-detection-methods.tex
\section{Current Action Detection Methods} \label{sec:current_methods}

Most action detection methods~\cite{Georgia-2015a,Weinzaepfel-2015,saha2017amtnet,Saha2016,peng2016eccv} follow an offline action tube generation approach.
These methods build \emph{action tubes} (i.e., sequences of detection bounding boxes around the action of interest, linked in time) by assuming that the entire video is available beforehand. 
Such methods are not suitable for self-driving cars because their action tube generation component is offline and slow, whereas for self-driving cars we want to process streaming videos online at real-time speed.
\\
Figure~\ref{fig:current_methods} illustrates an ideal self-driving system in action with an example.
A car is approaching a zebra crossing. 
The camera mounted on the car sends one or more streaming video sequences as input to an action detection module which processes, in real time, chunks of video, and outputs the class confidence and space-time locations (action tubes) of the various action and events it perceives.
Based on the action detection outputs, signals from other sensors mounted on the car and outputs from other modules (such as path planning, sign and object detection, self-localisation) an \emph{autonomous driving system} sends control signals to the car.
Note that, as far as the action detection system is concerned, \textbf{(1)} action tube generation has to be \textbf{online}: in the first instance action tubes need to be built based on the initial video chunk, whereas tubes are later incrementally updated in time as more and more chunks arrive. Furthermore \textbf{(2)}, the processing needs to take place in \textbf{real-time} to allow the driving system to swiftly react to new developments in its environment.

\begin{figure*}[t]
  \centering
  \includegraphics[scale=0.27]{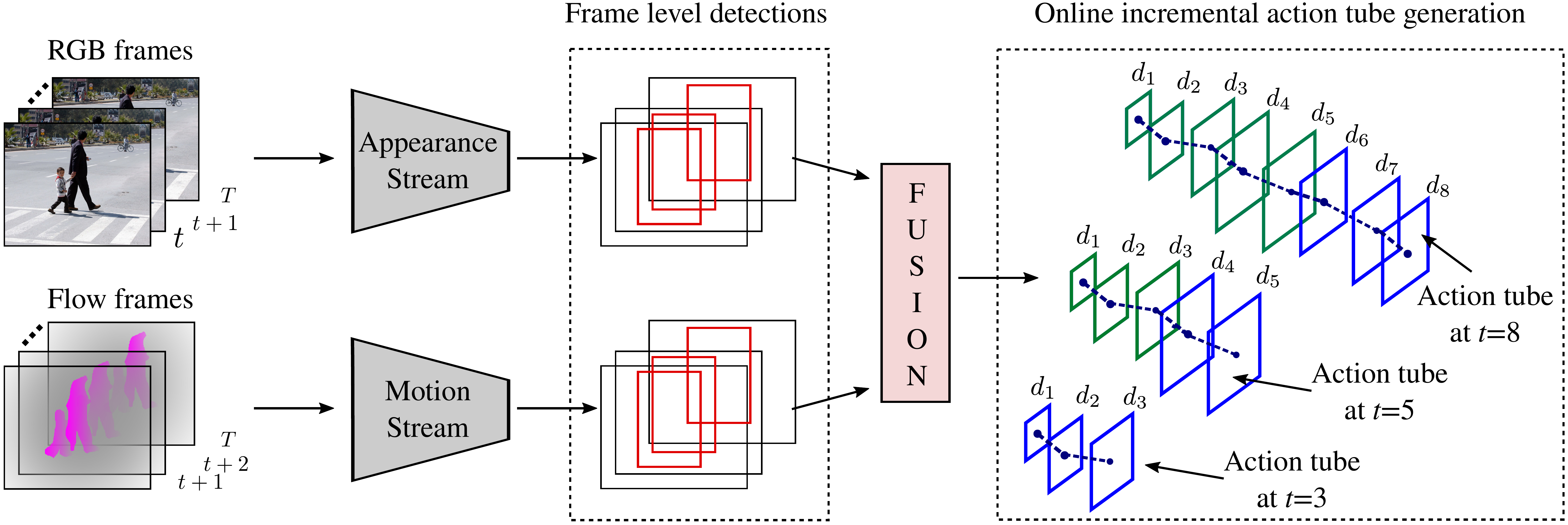}
%   \vskip 2mm
  \caption{{\small \textit{Online real-time action detection pipeline proposed by Singh~\etal~\cite{singh2016online}.}}}
 
\label{fig:guru_iccv2017_model} 
% \vspace{-0.6cm}
\end{figure*}

Most recently, Singh~\etal~\cite{singh2016online} have proposed an action tube generation algorithm which incrementally builds action tubes in an online fashion and at real-time speed, and is thus suitable for our task.
Following Singh~\etal~\cite{singh2016online}'s work, other authors \cite{kalogeiton2017action,ava2017gu} have used Singh~\etal's online action tube building algorithm, without though exhibiting real-time processing speed.
In the following section we briefly recall the online real-time action detection approach of \cite{singh2016online}, which is used in this work to report action detection results on our new Road Event and Activity Detection dataset.

\subsection{Online real-time action detection}

Figure~\ref{fig:guru_iccv2017_model} shows the block diagram of the online real-time action detection pipeline proposed by Singh~\etal~\cite{singh2016online}.
The pipeline takes RGB and optical flow frames as inputs and processes them through their respective appearance and motion streams.
The appearance and motion streams output frame level detection bounding boxes and their associated softmax scores are then fused using a late fusion scheme, which allows the system to
exploit the complementary aspects of appearance and motion information concerning the actions present in the video.
The resulting frame level detections are incrementally linked in time to build action tubes in an online fashion.
Unlike previous approaches~\cite{Georgia-2015a,Weinzaepfel-2015,saha2017amtnet,Saha2016,peng2016eccv}, Singh~\etal~\cite{singh2016online} use a faster and more efficient SSD fully convolutional neural network architecture~\cite{liu15ssd} to implement the appearance and motion streams.
Furthermore, \cite{singh2016online} proposes an elegant online action tube generation algorithm as opposed to the offline algorithms used by previous authors~\cite{Georgia-2015a,Weinzaepfel-2015,saha2017amtnet,Saha2016,peng2016eccv}.
Rather than generating action tubes using a Viterbi forward and backward pass (thus assuming that frame level detections are available for the entire video),
\cite{singh2016online} incrementally builds action tubes using only a Viterbi forward pass, starting by processing a smaller video snippet (a few initial frames) and incrementally updating the tubes as more and more frames are available to the system.

% ~\cite{Georgia-2015a,Weinzaepfel-2015,saha2017amtnet,kalogeiton2017action,hou2017tube,Saha2016,peng2016eccv,singh2016online}

% There are multiple action detection methods but only a few of them can perform action detection in online and real-time fashion.
% Now, we present real-time method present by Singh \etal for action detection. 
% We use this method as baseline and build on it to incorprate more models like AMTnet and TraMnet for online action detection.

\iffalse
but rather say we provde the inputs to self driving module
where fusion should happen from other sensors and other models 
we only doing part of the selfdriving moudle
we increasing the awarness level of such a module rather than whole module itself
becasue selfdriving module would contain path planing, sign detection, action detection, object detection self loclisation etc.
\fi

%% file: text/experiments.tex
\section{Experiments} \label{sec:experiment} % Fabio: pretty confused ...

In this section, we present action detection results on the initial version of the READ dataset.
We term this version READv1, which was created by annotating six days of videos from~\cite{maddern20171}.
Each day video is divided into multiple videos, usually 20-40 minutes long. 
%We annotated the agents with there actions and locations w.r.t to the AV. 
We followed the annotation process described in Section 3.
However, the resulting dataset had a long tail, % Fabio: what does this mean?
and some actions did not have many instances. 
We combined agents and actions to form event categories. 
If we consider location labels, then number final event classes increases hence the number of instances gets divided among these classes. 
As explained in Section~\ref{ssc:event_labels}, we picked 32 event classes as shown in Table~\ref{table:results} along with their number of instances. 
In this case, an event instance is an annotation of that particular event in a frame with a bounding box, and one frame can contain multiple instances of an event. 

The READv1 dataset contains 11K annotated frames in total: 4343 frames of them are used as test set, with the remaining ones used for training. 
These 11K frames are sampled from a broader set of frames coming from videos captured over 6 days, at the rate of 4 frames per second. 

\subsection{Detector details}

In the initial tests shown here, we train only the appearance stream of Singh~\etal~\cite{singh2016online}'s action detector (\S~\ref{sec:current_methods}) using READv1's annotations. 
We train the action detection network for 30K iterations with an initial learning rate of 0.001, up to 40K iterations. 
We plan to add the training of the flow stream and the fusion strategy when running tests on the next version of the dataset, which we plan to release by October 2018.

\subsection{Evaluation metric}

In these tests evaluation is done on a frame-by-frame basis, rather than on the basis of action tube detections. Namely, we use frame-AP~\cite{Georgia-2015a} as the evaluation metric, rather than video-mAP~\cite{Georgia-2015a} as the temporal association of ground truth bounding boxes are not yet fully available.
Nevertheless, we plan to make it available soon, in order to be able to evaluate this or any other model using video-mAP, which is the accepted, standard evaluation metric for action detection.
% the current version of the dataset has some errors in the ground truth linking process. 
% We plan to fix these errors and switch video-mAP as a standard evaluation metric. 
As is standard practice, we computed frame-AP results at a detection threshold ($\delta$, measuring the Intersection-over-Union (IOU) degree of overlap between ground truth and predicted action bounding box) equal to or greater than 0.5. 
%As the detection criterion, the  between detection and ground truth bounding boxes is used with the detection threshold and the correct ground-truth label. 

\subsection{Discussion}

Table~\ref{table:results} shows the action detection results on the test set of READv1. 
We can see a clear correlation between the number of instances (in the second column) and the performance of each class (in the third column). 
It indicates that an increase in the number of instances per class should improve detection performance. 
The final performance, in terms of frame-mAP, is a modest $17.5\%$, as the number of classes is limited.
\\
We are working to improve the dataset in a number of ways: 
i) by providing annotation in a multilabel format, as described in Section 3, to describe the different aspects of road event and activities;
ii) by annotating additional instances for the classes which have fewer number of instances, to avoid imbalance;
iii) by providing the temporal linking of detections across frames.

\begin{table}[h]
    \caption{Frame-mAP @ IoU 0.5 for event detection on the 4343 frames of test set.}
    \vspace{1mm}
    \resizebox{0.48\textwidth}{!}{ 
    \begin{tabular}{lcc}
        \toprule
        Event label & instances & AP@0.5 \\
        \midrule
        Car indicating right & 51 & 20.1\\
        Car turn right & 235 & 12.7\\
        Car indicating left & 1 & 00.0\\
        Car turn left & 51 & 18.7\\
        Car stopped at the traffic light & 1638 & 06.1\\
        Car moving in lane & 3098 & 45.8\\
        Car braking in lane & 102 & 04.7\\
        Car stopped in lane & 1404 & 61.9\\
        Car waiting at junction & 6 & 00.0\\
        edestrian waiting to cross & 452 & 03.8\\
        Pedestrian crossing road legally & 468 & 18.0\\
        Pedestrian walking on left pavement & 1930 & 31.1\\
        Pedestrian walking on right pavement & 1930 & 30.9\\
        Pedestrian walking on road left side & 48 &01.8\\
        Pedestrian walking on road right side & 48 &02.4\\
        Pedestrian crossing road illegally & 16 & 00.0\\
        Cyclist indicating left & 16 & 01.8\\
        Cyclist indicating right & 16 & 00.0\\
        Cyclist moving to left lane & 17 & 06.9\\
        Cyclist moving to right lane & 17 & 00.9\\
        Cyclist stopped at the traffic light & 790 & 39.8\\
        Cyclist turn right & 56 & 00.1\\
        Cyclist turn left & 56 & 00.0\\
        Cyclist crossing & 58 & 23.9\\
        Cyclist moving in lane & 2306 & 44.5\\
        Cyclist stopped in lane & 16 & 00.0\\
        Cyclist moving on pavement & 40 & 00.0\\
        Motorbike moving in lane & 1 & 00.0\\
        Bus moving in lane & 456 & 38.3\\
        Trafficlight red & 1526 & 52.7\\
        Trafficlight amber & 324 & 32.7\\
        Trafficlight green & 852 & 58.6\\
        \midrule
        total & 18027 & mAp=17.5\\
        \bottomrule
    \end{tabular}
    }

    \label{table:results}
\end{table}

%% file: text/conclusion.tex
\section{Conclusions} \label{sec:conclusions}

In this report we presented a new Road Event and Activity Detection (READ) dataset, as the first benchmark for road event detection in autonomous driving. READ has been constructed by providing extra annotation to a fraction of the recently released Oxford RobotCar dataset. The annotation provided follows a multi-label approach in which road agents (including the AV), their locations and the action they perform (possibly more than one) are labelled separately. Event-level labels can be generated by simply composing lower-level descriptions.

Here we showed preliminary tests conducted using the current state of the art in online action detection, using only frame-mAP as a metric, and on a small subset of the final dataset, using an initial set of event labels. In the upcoming months, prior to release of the full dataset, we will work towards (i) completing the multi-label annotation of around 40,000 frames coming from videos spanning a wide range of road conditions; (ii) providing the temporal association ground truth information necessary to compute video-mAP results; (iii) devising a novel, deep learning approach to detecting complex activities, such as those associated with common driving situations. All data, documentation and baseline code will be publicly released on GitHub.